# Image Synthesis in Multi-Contrast MRI with Conditional Generative Adversarial Networks

Salman Ul Hassan Dar[1,2], Mahmut Yurt[1,2], Levent Karacan[4], Aykut Erdem[4], Erkut Erdem[4], Tolga Çukur[1,2,3]

[1]Department of Electrical and Electronics Engineering, Bilkent University, Ankara, Turkey
[2]National Magnetic Resonance Research Center (UMRAM), Bilkent University, Ankara, Turkey
[3]Neuroscience Program, Sabuncu Brain Research Center, Bilkent University, Ankara, Turkey
[4]Computer Engineering, Hacettepe University, Ankara, Turkey

*Running title:* Image Synthesis in Multi-Contrast MRI with Conditional GANs.

*Address correspondence to:*

Tolga Çukur

Department of Electrical and Electronics Engineering, Room 304
Bilkent University
Ankara, TR-06800, Turkey

TEL: +90 (312) 290-1164

E-MAIL: cukur@ee.bilkent.edu.tr

This work was supported in part by a European Molecular Biology Organization Installation Grant (IG 3028), by a TUBA GEBIP fellowship, and by a BAGEP fellowship awarded to T. Çukur. We gratefully acknowledge the support of NVIDIA Corporation with the donation of the Titan GPUs used in this study.

To be submitted to IEEE Transactions on Medical Imaging

**Abstract**

Acquiring images of the same anatomy with multiple different contrasts increases the diversity of diagnostic information available in an MR exam. Yet, scan time limitations may prohibit acquisition of certain contrasts, and images for some contrast may be corrupted by noise and artifacts. In such cases, the ability to synthesize unacquired or corrupted contrasts from remaining contrasts can improve diagnostic utility. For multi-contrast synthesis, current methods learn a nonlinear intensity transformation between the source and target images, either via nonlinear regression or deterministic neural networks. These methods can in turn suffer from loss of high-spatial-frequency information in synthesized images. Here we propose a new approach for multi-contrast MRI synthesis based on conditional generative adversarial networks. The proposed approach preserves high-frequency details via an adversarial loss; and it offers enhanced synthesis performance via a pixel-wise loss for registered multi-contrast images and a cycle-consistency loss for unregistered images. Information from neighboring cross-sections are utilized to further improved synthesis quality. Demonstrations on $T_1$- and $T_2$-weighted images from healthy subjects and patients clearly indicate the superior performance of the proposed approach compared to previous state-of-the-art methods. Our synthesis approach can help improve quality and versatility of multi-contrast MRI exams without the need for prolonged examinations.

## 1 - Introduction

Magnetic resonance imaging (MRI) is pervasively used in clinical applications due to the diversity of contrasts it can capture in soft tissues. Tailored MRI pulse sequences enable the generation of distinct contrasts while imaging the same anatomy. For instance, $T_1$-weighted brain images clearly delineate gray and white matter tissues, whereas $T_2$-weighted images delineate fluid from cortical tissue. In turn, multi-contrast images acquired in the same subject increase the diagnostic information available in clinical and research studies. However, it may not be possible to collect a full array of contrasts given considerations related to the cost of prolonged exams and uncooperative patients. In such cases, acquisition of contrasts with relatively shorter scan times might be preferred. Moreover, in cohort studies, a subset of the acquired contrasts is typically corrupted by excessive noise or artifacts in some subjects that prohibit subsequent diagnostic use [1]. Therefore, the ability to retrospectively synthesize missing or corrupted contrasts from other successfully acquired contrasts has potential value for enhancing multi-contrast MRI, and improve analysis tasks such as registration and segmentation [2].

Cross-domain synthesis of medical images have recently been gaining popularity in medical imaging. Given a subject's image $b_1$ in $M_1$ (source domain), the aim is to accurately estimate the respective image of the same subject $b_2$ in $M_2$ (target domain). Depending on how they tackle this problem, there are two main approaches: registration-based and intensity-transformation-based methods [3]. Registration-based methods start by generating an atlas based on a co-registered set of images, $a_1$ and $a_2$, respectively acquired in $M_1$ and $M_2$ [4]. These methods further make the assumption that within-domain images from separate subjects are related to each other through a geometric warp. For synthesizing $b_2$ from $b_1$, the warp that transforms $a_1$ to $b_1$ is estimated, and this warp is then applied on $a_2$. Since they only rely on geometric transformations, registration-based methods suffer from across-subject differences in underlying morphology [3]. For example, inconsistent pathology across a test subject and the atlas can cause failure. Furthermore, within-domain registration accuracy might be limited even in normal subjects.

An alternative is to use intensity-based methods that do not rely on a strict geometric relationship among different subjects' anatomies [3], [5]–[8]. One powerful approach for multi-contrast MRI is based on the compressed sensing framework, where each patch in the source image $b_1$ is expressed as a sparse linear combination of patches in the atlas image $a_1$ [7]. The learned sparse combinations are then applied to estimate patches in $b_2$ from patches in $a_2$. To improve matching of patches across domains, generative models were also proposed that use multi-scale patches and tissue segmentation labels [9], [10]. Instead of focusing on linear models, recent studies aimed to learn more general non-linear mappings that express individual voxels in $a_2$ in terms of patches in $a_1$, and then predict $b_2$ from $b_1$ based on these mappings. Nonlinear mappings are learned on training data via techniques such as nonlinear regression [3], [5], [6] or location-sensitive neural networks [11]. An important example is Replica that performs random forest regression on multiresolution image patches [3]. Replica demonstrates great promise in multi-contrast MR image synthesis. However, patches at different spatial scales are treated independently during dictionary construction, and the predictions from separate random forest trees are averaged during synthesis. These may lead to loss of high spatial frequency information and suboptimal synthesis performance.

Recently an end-to-end framework for MRI image synthesis has been proposed, Multimodal, based on deep neural networks [12]. Multimodal trains a neural network that receives as input images in several different source contrasts, and predicts the image in the target contrast. This method performs multiresolution dictionary construction and image synthesis in a unified framework, and it was demonstrated to yield higher synthesis quality compared to non-network-based approaches even when only a subset of the source contrasts is available. That said, Multimodal assumes the availability of spatially-registered multi-contrast images. In addition, Multimodal uses mean squared or absolute error loss functions that can perform poorly in capturing errors at high spatial frequencies [13]–[15].

Here we propose a novel approach for image synthesis in multi-contrast MRI based on generative adversarial network (GAN) architectures. Adversarial loss functions have recently been demonstrated for image-to-image translation with reliable capture of high-frequency texture information [16], [17]. Inspired by this success, our proposed method employs conditional GANs to synthesize images in the target contrast given input images in the source contrast. For improved accuracy, the proposed method leverages correlated information across neighboring cross-sections within a volume during synthesis. Distinct implementations are used for two different scenarios, where multi-contrast images are spatially registered (pGAN) and where they are unregistered (cGAN). For the first scenario, we train pGAN with pixel-wise loss between the synthesized and true images [16]. For the second scenario, we train cGAN after replacing the pixel-wise loss with a cycle loss that enforces the ability to reconstruct back the source image from the synthesized target image [17]. Extensive evaluations are presented on multi-contrast MRI images ($T_1$- and $T_2$-weighted) from healthy normals and glioma patients. The proposed approach yields visually and quantitatively enhanced accuracy in multi-contrast MRI synthesis compared to state-of-the-art methods.

A summary of our main contributions in this study are listed below:

1. We demonstrate a novel end-to-end image synthesis approach for MRI that successfully estimated the image in the target contrast given the image in the source contrast.
2. To our knowledge, this is the first method that utilizes conditional GANs with pixel-wise and cycle-consistency loss functions for multi-contrast MRI synthesis.
3. To our knowledge, this is the first method that can perform multi-contrast MRI synthesis based on unregistered images.
4. We show that aggregating information across neighboring cross-sections within the source volume enables enhanced synthesis of the target MR images.

## 2 - Methods

### 2.1 – Image synthesis via adversarial networks

Generative adversarial networks are neural-network architectures that consist of two sub-networks; $G$, a generator and $D$, a discriminator. $G$ learns a mapping from a latent variable $z$ (typically random noise) to an image $y$ in a target domain, and $D$ learns to discriminate the generated image $G(y)$ from the real image $y$ [18]. During training of a GAN, both $G$ and $D$ are learned simultaneously, with $G$ aiming to generate images that are indistinguishable from the real images, and $D$ aiming to tell apart generated and real images. To do this, the following adversarial loss function ($\mathcal{L}_{GAN}$) can be used:

$$\mathcal{L}_{GAN}(G,D) = E_y[logD(y)] + E_z[\log(1 - D(G(z)))] \quad (1)$$

where $G$ tries to minimize and $D$ tries to maximize the adversarial loss. The adversarial loss helps the generator in modeling high-spatial-frequency information [14]. Both the generator and the discriminator are trained simultaneously. The desired behavior upon convergence is that $G$ is capable of producing realistic counterfeit images that $D$ cannot recognize [18].

Recent studies in computer vision have demonstrated that GANs are very effective in image-to-image translation tasks [16], [17]. Image-to-image translation concerns transformations between different representations of the same underlying visual scene [16]. These transformations can be used to convert an image between separate domains, e.g., generating semantic segmentation maps from images, colored images from sketches, or maps from aerial photos [16], [19], [20]. Traditional GANs learn to generate samples of images from noise. However, in image-to-image translation, the synthesized image has statistical dependence on the source image. To better capture this dependency, conditional GANs can be employed that receive the source image as an additional input [21]. The resulting network can then be trained based on the following adversarial loss function:

$$\mathcal{L}_{condGAN}(G,D) = E_{x,y}[logD(x,y)] + E_{x,z}[\log(1 - D(x,G(x,z)))] \quad (2)$$

where $x$ denotes the source image.

An analogous problem image-to-image translation tasks in computer vision exists in MR imaging where the same anatomy is acquired under multiple different tissue contrasts (e.g., $T_1$- and $T_2$-weighted images). Inspired by the recent success of adversarial networks, here we employed conditional GANs to synthesize MR images of a target contrast given as input an alternate contrast. For a comprehensive solution, we considered two distinct scenarios for multi-contrast MR image synthesis. First, we assumed that the images of the source and target contrasts are perfectly registered. For this scenario, we propose pGAN that incorporates a pixel-wise loss into the objective function as inspired by the pix2pix architecture [16]:

$$\mathcal{L}_{L1}(G) = E_{x,y,z}[\|y - G(x,z)\|_1] \quad (3)$$

where $\mathcal{L}_{L1}$ is the pixel-wise L1 loss function. Since the generator $G$ was observed to ignore the latent variable in pGAN, the following aggregate loss function was used for training:

$$\mathcal{L}_{pGAN} = \mathcal{L}_{condGAN}(G,D) + \lambda \mathcal{L}_{L1}(G) \quad (4)$$

where $\mathcal{L}_{pGAN}$ is the complete loss function, and $\lambda$ is the parameter that controls the relative weighing of the pixel-wise loss.

In the second scenario, we did not assume any explicit registration between the images of the source and target contrasts. In this more realistic scenario, the pixel-wise loss cannot be leveraged since pixels are not aligned between contrasts. To limit the number of potential solutions for the synthesized image, here we proposed cGAN that incorporates a cycle-consistency loss as inspired by the cycleGAN architecture [17]. The cGAN method consists of two generators ($G_x, G_y$) and two discriminators ($D_x, D_y$). $G_y$ tries to generate $G_y(x)$ that looks similar to $y$ and $D_y$ tries to distinguish $G_y(x)$ from the images $y$. On the other hand, $G_x$ tries to generate $G_x(y)$ that looks similar to $x$ and $D_x$ tries to distinguish $G_x(y)$ from the images $x$. This architecture incorporates an additional loss to ensure that the input and target images are consistent with each other, called as the cycle consistency loss $\mathcal{L}_{cycle}$:

$$\mathcal{L}_{cycle}(G_x, G_y) = E_x\left[\|x - G_x(G_y(x))\|_1\right] + E_y\left[\|y - G_y(G_x(y))\|_1\right] \quad (5)$$

This loss function enforces that property that after projecting the source images onto the target domain, the source image can be re-synthesized with minimal loss from the projection. Further replacing the negative log-likelihood cost for adversarial loss in Eq. 1 by a squared loss [22]:

$$\mathcal{L}_{GAN}(G, D) = E_y[(D(y) - 1)^2] + E_x[D(G(x))^2] \quad (6)$$

yields the following aggregate loss function for training:

$$\mathcal{L}_{cGAN}(D_x, D_x, G_x, G_y) = \mathcal{L}_{GAN}(G_x, D_x) + \mathcal{L}_{GAN}(G_y, D_y) + \lambda \mathcal{L}_{cycle}(G_x, G_y). \quad (7)$$

**2.2 – MRI datasets**

For registered images, we trained and test both pGAN and cGAN models. For unregistered images, we only trained cGAN models. The experiments were performed on three separate datasets: the MIDAS dataset [23], the IXI dataset (*http://brain-development.org/ixi-dataset/*) and the BRATS dataset (*https://sites.google.com/site/braintumorsegmentation/home/brats2015*). MIDAS and IXI datasets primarily contained data from healthy subjects, whereas the BRATS datasets contained data from patients with structural abnormality (i.e., brain tumor). Protocol information on images included in each dataset are described below. Datasets were normalized to ensure comparable ranges of voxel intensities across subjects. For each contrast, the mean intensity across the brain volume was normalized to 1 within individual subjects. To attain an intensity scale in [0 1], three standard deviations above the mean intensity of voxels pooled across subjects was then mapped to 1.

**2.2.1 - MIDAS dataset**
Data from 66 subjects were analyzed. A total of 4865 $T_1$-weighted images were assembled, where 3774 were used for training and 1091 were reserved for testing. A total of 4865 $T_2$-weighted images were assembled, where 3774 were used for training and 1091 were reserved for testing. The $T_1$- and $T_2$-weighted images analyzed here were acquired via the following parameters. $T_1$-weighted images: 3D gradient-echo sequence, TR=14ms, TE=7.7ms, flip angle=$25^0$, matrix size=256x176, 1 mm isotropic resolution, axial orientation. $T_2$-weighted images: 2D spin-echo sequence, TR=7730ms, TE=80ms, flip angle=$180^0$, matrix size=256x192, 1 mm isotropic resolution, axial orientation.

### 2.2.2 - IXI dataset

Data from 30 subjects were analyzed. A total of 3320 $T_1$-weighted images were assembled, where 2780 were used for training and 540 were reserved for testing. A total of 2730 $T_2$-weighted images were assembled, where 2275 were used for training and 455 were reserved for testing. The $T_1$- and $T_2$-weighted images analyzed here were acquired with the following parameters. $T_1$-weighted images: TR=9.813ms, TE=4.603ms, flip angle=$8^0$, volume size = 256 × 256 × 150, voxel dimensions = 0.94 × 0.94 × 1.2 mm$^3$, sagittal orientation. $T_2$-weighted images: TR=8178ms, TE=100ms, flip angle=$90^0$, volume size = 256 × 256 × 150, voxel dimensions = 0.94 × 0.94 × 1.2 mm$^3$, axial orientation.

### 2.2.3 - BRATS dataset

Data from 28 subjects with visible lesions were analyzed. A total of 2828 $T_1$-weighted images were assembled, where 2424 were used for training and 404 were reserved for testing, and a total of 2828 $T_2$-weighted images were assembled, where 2424 were used for training and 404 were reserved for testing. The BRATS dataset compiles data acquired under different scanning protocols employed on separate sites. Thus, a common protocol did not exist among $T_1$-weighted or $T_2$-weighted images analyzed here.

### 2.3 – Image registration

For the first scenario, multi-contrast images from a given subject were assumed to be registered. Note that the images contained in the MIDAS and IXI datasets are unregistered. Thus, the $T_1$- and $T_2$-weighted images in these datasets were registered prior to network training. For the MIDAS dataset, an affine transformation was used for registration based on a mutual-information cost. For the IXI datasets, a rigid transformation based on a mutual-information costs performed better. No registration was needed for the BRATS dataset. No registration was performed for the second scenario. All registrations were performed using FSL package [24], [25].

### 2.4 – Network training

Since we consider two different scenarios for multi-contrast MR image synthesis, network training procedures for these scenarios were distinct. In the first scenario, we assumed perfect alignment between the source and target images, and we then used pGAN to learn the mapping from the source to the target contrast. In a first variant of pGAN, the input image was a single cross-section of the source contrast, and the target was the respective cross-section of the desired contrast. Note that neighboring cross sections in MR images are expected to show significant correlation. Thus, we reasoned that incorporating additional information from adjacent cross-sections in the source contrast should improve synthesis. To do this, a second variant of pGAN was implemented where multiple consecutive cross-sections of the source contrast were given as input, with the target corresponding to desired contrast at the central cross-section. It was observed that using three cross-sections yielded nearly optimal results, without substantially increasing the model complexity. Thus, implementations of the second variant were based on three cross-sections thereafter.

The implementation of the pGAN network provided in *https://github.com/junyanz/pytorch-CycleGAN-and-pix2pix* was used based on 256x256 images. Thus all MR images were zero-padded in the image-domain to this size prior to training. We adopted the generator architecture from [26], and the discriminator architecture from [13]. The training procedures lasted for 200 epochs, and the Adam optimizer was used with a minibatch size of 1 [27]. In the first 100 epochs, the learning rate for both the generator and the discriminator was fixed at 0.0002. In the remaining 100 epochs, the learning rate was linearly decayed from 0.0002 to 0. Decay rates for the first and second moments of gradient estimates were set as β1= 0.5 and β2=0.999, respectively. Instance normalization was also applied [28]. All weights were initialized using normal distribution with 0 mean and 0.02 std.

In the second scenario, we did not assume any alignment between the source and target images, and so we used cGAN to learn the mapping between unregistered source and target images. Similar to pGAN, two variants of cGAN were considered that worked on a single cross-section and on three consecutive cross-sections. The latter variant of cGAN was implemented where multiple consecutive cross-sections of the source contrast were given as input, and an equal number of consecutive cross-sections of the target contrast were taken as output. Although cGAN does not assume any alignment between the source and target domains, we still wanted to quantitatively examine the effects of distinct loss functions used in cGAN and pGAN. For comparison purposes, we also trained separate cGAN networks on registered multi-contrast data ($cGAN_{reg}$). The training procedures were identical to those for pGAN.

## 2.5 – Competing methods

To comparatively demonstrate the proposed approach, two state-of-the-art methods for MRI image synthesis were implemented. The first method was Replica that estimates a nonlinear mapping between image patches from the source contrast onto individual voxels from the target contrast [3]. Replica extracts image features at different spatial scales, and then performs a multi-resolution analysis via random forests. The learned nonlinear mapping is then applied on test images. Code posted by the authors of the Replica method was used to train the models, based on the parameters described in [3].

The second method was Multimodal that uses an end-to-end neural network to estimate the target image given the source image as input. A neural-network implementation implicitly performs multi-resolution feature extraction and synthesis based on these features. Trained networks can then be applied on test images. Code posted by the authors of the Multimodal method was used to train the networks, based on the parameters described in [12].

For comparisons between the proposed approach and the competing methods, the same set of training and test data were used. Since the proposed models were implemented for unimodal mapping between two separate contrasts, Replica and Multimodal implementations were also performed with only two contrasts.

## 2.6 - Experiments

Here we first questioned whether the direction of registration between multi-contrast images affects the quality of synthesis. In particular, we generated multiple registered datasets from $T_1$- and $T_2$-weighted images. In the first set, $T_2$-weighted images were registered onto $T_1$-weighted images (yielding $T_2$*). In the second set, $T_1$-weighted images were registered onto $T_2$-weighted images (yielding $T_1$*). In addition to the direction of registration, we also considered the two possible directions of synthesis ($T_2$ from $T_1$; $T_1$ from $T_2$).

For originally unregistered datasets (i.e., MIDAS and IXI), the above-mentioned considerations led to four distinct cases: a) $T_1 \rightarrow T_2$*, b) $T_1$* $\rightarrow T_2$, c) $T_2 \rightarrow T_1$*, d) $T_2$* $\rightarrow T_1$. Here, $T_1$ and $T_2$ are unregistered images, $T_1$* and $T_2$* are registered images, and $\rightarrow$ corresponds to the direction of synthesis. For each case, pGAN and cGAN networks were trained based on two variants, one receiving a single cross-section as input, the other receiving three consecutive cross-sections as input. This resulted in a total of 8 pGAN and 4 cGAN models. Note that a single cGAN architecture contains generators for both contrasts and trains a model that can synthesize in both directions. For readily registered datasets (i.e., BRATS), no registration was needed and this resulted in only two distinct cases for consideration: a) $T_1 \rightarrow T_2^*$ and d) $T_2^* \rightarrow T_1$. Two variants of pGAN and cGAN were considered that work on a single cross-section and three cross-sections.

To investigate how well the proposed models perform with respect to state-of-the-art approaches, we compared the performance of pGAN and cGAN models with two previous the Replica and Multimodal methods. Since the adversarial loss in the GAN architecture enforces synthesis of realistic images, we predicted that pGAN and cGAN would outperform these competing methods.

Models under comparison were trained and tested on the same data. The synthesized images obtained from each model were compared with the true target images as reference. Both the synthesized and the reference images were normalized to a maximum intensity of 1. To assess the synthesis quality, we measured the peak signal-to-noise ratio (PSNR) and structural similarity index (SSIM) metrics between the synthesized image and the reference. Neural network training and evaluation was performed on NVIDIA Titan X Pascal and Xp GPUs. Implementation of pGAN and cGAN were carried out in Python using the Pytorch framework [29]. A MATLAB implementation of REPLICA, and a Keras implementation [30] of Multimodal with the Theano backend [31] were used.

# 3 – Results

We first evaluated the proposed conditional GAN architectures on $T_1$-and $T_2$-weighted images from the MIDAS dataset. For evaluation, we considered two cases for $T_2$ synthesis (a. $T_1 \rightarrow T_2^*$, b. $T_1^* \rightarrow T_2$, where * denotes the registered image), and two cases for $T_1$ synthesis (c. $T_2 \rightarrow T_1^*$, d. $T_2^* \rightarrow T_1$). Table I lists the PSNR and SSIM measurements across test images for all cases, based on pGAN, cGAN$_{reg}$ trained on registered data, and cGAN$_{unreg}$ trained on unregistered data. We find that pGAN outperforms both cGAN$_{unreg}$ and cGAN$_{reg}$ in all cases, except in $T_2^* \rightarrow T_1$ where cGAN$_{reg}$ performs similarly. In terms of PSNR, pGAN outperforms its closest competitor cGAN$_{reg}$ by 2.72 dB in $T_2$ synthesis, and by 0.55 dB in $T_1$ synthesis (on average across two cases). These improvements can be attributed to the benefits of the pixel-wise loss compared to the cycle-consistency loss on paired, registered images. Representative results for $T_2$ synthesis ($T_1 \rightarrow T_2^*$) and $T_1$ synthesis ($T_2^* \rightarrow T_1$) are displayed in Figs. 3a and 4a, respectively. The pGAN method yields higher synthesis quality compared to cGAN$_{reg}$. Although cGAN$_{unreg}$ was trained on unregistered images, it can faithfully capture fine-grained structure in the synthesized contrast. Overall, both pGAN and cGAN yield synthetic images of remarkable visual similarity to the reference.

The source contrast images that are input to the networks can occasionally contain considerable levels of noise. In such cases, synthesis quality might be improved by incorporating correlated structural information across neighboring cross-sections. To examine this issue, we trained pGAN, cGAN$_{reg}$ and cGAN$_{unreg}$ with three consecutive cross-sections in the source domain as input (see Methods). Table II lists the PSNR and SSIM measurements across test images for $T_2$ and $T_1$ synthesis, considering the two potential directions of registration in each synthesis task. Representative results for $T_2$ synthesis ($T_1 \rightarrow T_2^*$) and $T_1$ synthesis ($T_2^* \rightarrow T_1$) using multi cross-section models are as shown in Figs. 3b and 4b. As expected for pGAN, the multi cross-section model outperforms the single cross-section model by 0.55 dB in $T_2$, and by 0.99 dB in $T_1$ synthesis in terms of PSNR. These quantitative measurements are also affirmed by improvements in visual quality for the multi cross-section model in Figs. 3 and 4. In contrast, the benefits are less clear for cGAN. Note that, unlike pGAN that works on paired images, the discriminators in cGAN work on unpaired images from the source and target domains. In turn, this can render incorporation of correlated information across cross sections less effective.

Next, we comparatively demonstrated the proposed methods against two state-of-the-art techniques for multi-contrast MRI synthesis, namely Replica and Multimodal. For this purpose, we trained and tested pGAN, cGAN$_{reg}$, Replica, and Multimodal on $T_1$- and $T_2$-weighted brain images from the IXI dataset. Note that Replica performs ensemble averaging across random forest trees and Multimodal uses mean-squared or absolute error measures that may lead to overemphasis of low frequency information. In contrast, conditional GANs leverage loss functions that can more effectively capture details at high spatial frequencies. Thus, we predicted that pGAN and cGAN would synthesize sharper and more realistic images as compared to the competing methods. Table III lists the PSNR and SSIM measurements across test images synthesized via pGAN, cGAN$_{reg}$, Replica and Multimodal. As predicted, both pGAN and cGAN outperform the competing methods in terms of PSNR in all examined cases, where pGAN is the top-performing method. On average, pGAN achieves 1.04 dB higher PSNR than Multimodal in $T_2$ synthesis, and 2.41 dB higher PSNR in $T_1$ synthesis. Compared to Replica, pGAN improves PSNR by 2.89 dB in $T_2$ synthesis, and by 5.22 dB in $T_1$ synthesis. These performance differences are also visible in Figs. 5 and 6. The proposed pGAN method is superior in depiction high spatial frequency information compared to Replica and Multimodal.

Lastly, we demonstrated the performance of the proposed methods on images acquired from patients with clear pathology. To do this, we trained and tested pGAN, cGAN$_{reg}$, Replica, and Multimodal on $T_1$- and $T_2$-weighted brain images from the BRATS dataset. Similar to the previous evaluations, here we expected that the proposed method would synthesize more realistic images with improved preservation of fine-grained tissue structure. Table IV lists the PSNR and SSIM measurements across test images synthesized via pGAN, cGAN$_{reg}$, Replica and Multimodal. Again, the top-performing method is pGAN, and both pGAN and cGAN outperform competing methods. On average, pGAN achieves 2.34 dB higher PSNR than Multimodal in $T_2$ synthesis, and 2.27 dB higher PSNR in $T_1$ synthesis. Compared to Replica, pGAN increases PSNR by 1.65 dB in $T_2$ synthesis, and by 2.53 dB in $T_1$ synthesis. Representative images for $T_2$ and $T_1$ synthesis are displayed in Figs. 7 and 8, respectively. It is observed that pathologies present only in the source contrast can occasionally cause artefactual synthesis in Replica and Multimodal (see Fig. 8). Meanwhile, the pGAN method enables reliable synthesis with visibly improved depiction of high spatial frequency information.

## 4 – Discussion

A multi-contrast MRI synthesis approach based on conditional GANs was demonstrated against state-of-the-art methods in three publicly available brain MRI datasets (MIDAS, IXI, BRATS). The proposed pGAN method uses adversarial loss functions and correlated structure across neighboring cross-sections for improved synthesis, particularly at high spatial frequencies. While many previous methods require registered multi-contrast images for training, a cGAN method was presented that uses cycle-consistency loss for learning to synthesize from unregistered images. Comprehensive evaluations were performed for two distinct scenarios where training images were registered or unregistered within single subjects. In both scenarios, the proposed methods achieved higher synthesis quality both visually and in terms of quantitative assessments based on PSNR and SSIM metrics. While our synthesis approach was primarily demonstrated for multi-contrast brain MRI here, it may offer similar performance benefits in synthesis across imaging modalities such as MRI, CT or PET [32].

The conventional approach to image synthesis is to first construct a dictionary from multi-resolution image patches, then learn a sparse or nonlinear mapping between contrasts, and finally apply this mapping for synthesis of the target contrast. While recent dictionary-based methods have produced promising results [3], [5]–[8], the feature extraction stage to form the dictionary and the nonlinear mapping stage to learn the transformation between the source and target features are segregated. In contrast, the conditional GAN methods proposed here perform feature extraction and mapping simultaneously. Such end-to-end trained networks alleviate suboptimality due to independent optimization of separate synthesis stages. Here we observed improved spatial acuity in display of high spatial frequency information in synthesized brain images.

Several previous studies proposed the use of neural network architectures for multi-contrast MRI synthesis tasks [11], [12], [33]. A recent method, Multimodal, was demonstrated to yield higher quality synthesis compared to conventional methods in brain MRI datasets [12]. Unlike previous neural-network-based methods, the GAN architectures proposed here are generative networks that learn the conditional probability distribution of the target contrast given the source contrast. The incorporation of adversarial loss as opposed to typical squared or absolute error loss leads to enhanced capture of detailed texture information about the target contrast, thereby enabling higher synthesis quality.

The proposed synthesis approach might be further improved by considering several lines of technical development. Here we presented multi-contrast MRI synthesis results while considering two potential directions for image registration (e.g., $T_1 \rightarrow T_2^*$ and $T_1^* \rightarrow T_2$ for $T_2$ synthesis). We observed that the proposed methods yielded high-quality synthesis regardless of the registration direction. Note that comparisons between the two directions based on reference-based metrics are not informative because the references for the two directions are inevitably distinct (e.g., $T_2^*$ versus $T_2$), so determining the optimal direction of registration is challenging. Still, we expect that in cases of substantial mismatch between the voxel sizes in the source and target contrasts, the cGAN method will not only learn to synthesize but it will also attempt to learn to interpolate from the spatial sampling grid of the source onto that of the target. To alleviate potential performance loss, the spatial transformation between the source and target images can first be estimated via multi-modal registration. This interpolation function can then be cascaded to the output of the cGAN architecture.

Synthesis accuracy can also be improved by generalizing the current approach to predict the target based on multiple source contrasts. In principle, both pGAN and cGAN methods can be modified to receive as input multiple source contrasts in addition to multiple cross sections as demonstrated here.

In turn, this generalization can offer improved performance when a subset of the source contrast is unavailable. The performance of conditional GAN architectures in the face of missing inputs warrants further investigation. Alternatively, an initial fusion step could be incorporated that combines multi-contrast source images in the form of a single fused image fed as input to the GAN [34].

An important concern regarding neural-network based methods is the availability of large datasets for successful training of network parameters. The cGAN method facilitates network training by permitting the use of unregistered and unpaired multi-contrast datasets. While here we performed training on paired images for unbiased comparison, cGAN permits the use of unpaired images from distinct sets of subjects. As such, it can facilitate compilation of large datasets that would be required for improved performance via deeper networks. Yet, further performance improvements may be viable by training networks based on a mixture of paired and unpaired training data [35].

# 5 – Conclusion

We proposed a new multi-contrast MRI synthesis method based on conditional generative adversarial networks. Unlike most conventional methods, the proposed method performed end-to-end training of GANs that synthesize the target contrast given images of the source contrast. The use of adversarial loss functions improved accuracy in synthesis of high-spatial-frequency information in the target contrast. Synthesis performance was further improved by incorporating a pixel-wise loss in the case of registered images, and a cycle-consistency loss for unregistered images. Finally, the proposed method leveraged information across neighboring cross-sections within each volume to increase accuracy of synthesis. The proposed method outperformed two state-of-the-art synthesis methods in multi-contrast brain MRI datasets from healthy subjects and glioma patients. Therefore, our GAN-based approach holds great promise for multi-contrast synthesis in clinical practice.

**FIGURES**

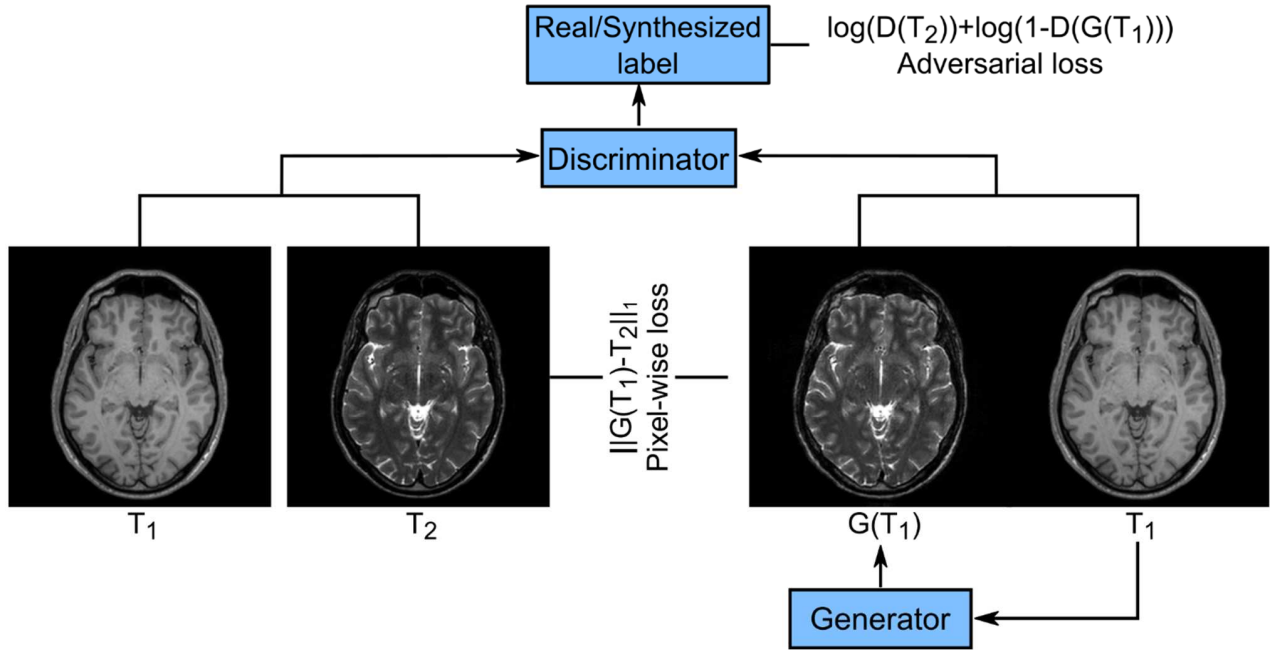

**Figure 1.** The pGAN method is based on a conditional adversarial network with a generator $G$ and a discriminator D. Given an input image in a source contrast (e.g., $T_1$-weighted), G learns to generate the image of the same anatomy in a target contrast (e.g., $T_2$-weighted). Meanwhile, D learns to discriminate between synthetic (e.g., $T_1$-$G(T_1)$) and real (e.g., $T_1$-$T_2$) pairs of multi-contrast images. Both subnetworks are trained simultaneously, where G aims to minimize a pixel-wise and an adversarial loss function, and D tries to maximize the adversarial loss function.

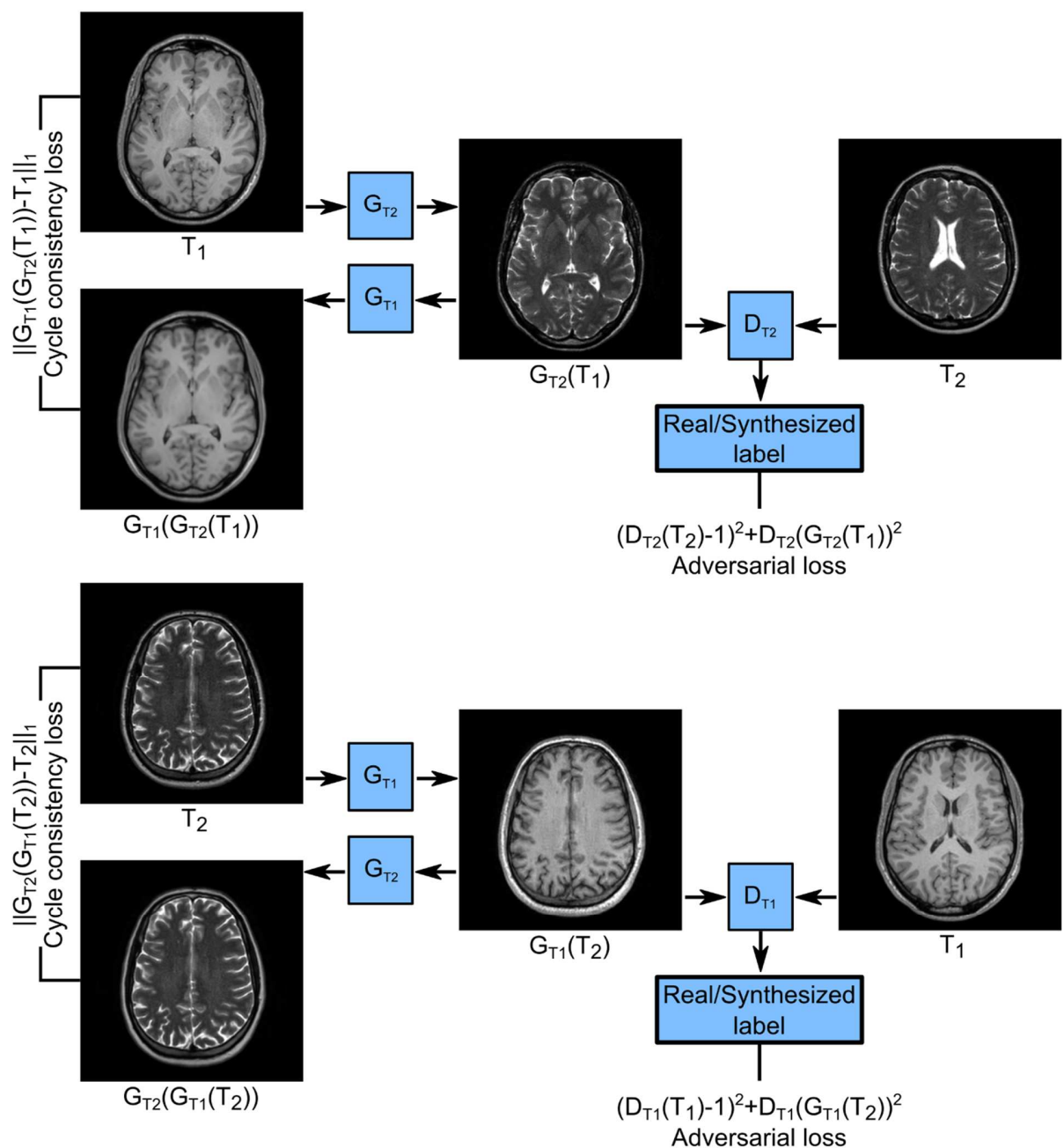

**Figure 2.** The cGAN method is based on a conditional adversarial network with two generators ($G_{T1}, G_{T2}$) and two discriminators ($D_{T1}, D_{T2}$). Given a T$_1$-weighted image, $G_{T2}$ learns to generate the respective T$_2$-weighted image of the same anatomy that is indiscriminable from real T$_2$-weighted images of other anatomies, whereas $D_{T2}$ learns to discriminate between synthetic and real T$_2$-weighted images. Similarly, $G_{T1}$ learns to generate realistic a T$_1$-weighted image of an anatomy given the respective T$_2$-weighted image, whereas $D_{T1}$ learns to discriminate between synthetic and real T$_1$-weighted images. Since the discriminators do not compare target images of the same anatomy, a pixel-wise loss cannot be used. Instead, a cycle-consistency loss is utilized to ensure that the trained generators enable reliable recovery of the source image from the generated target image.

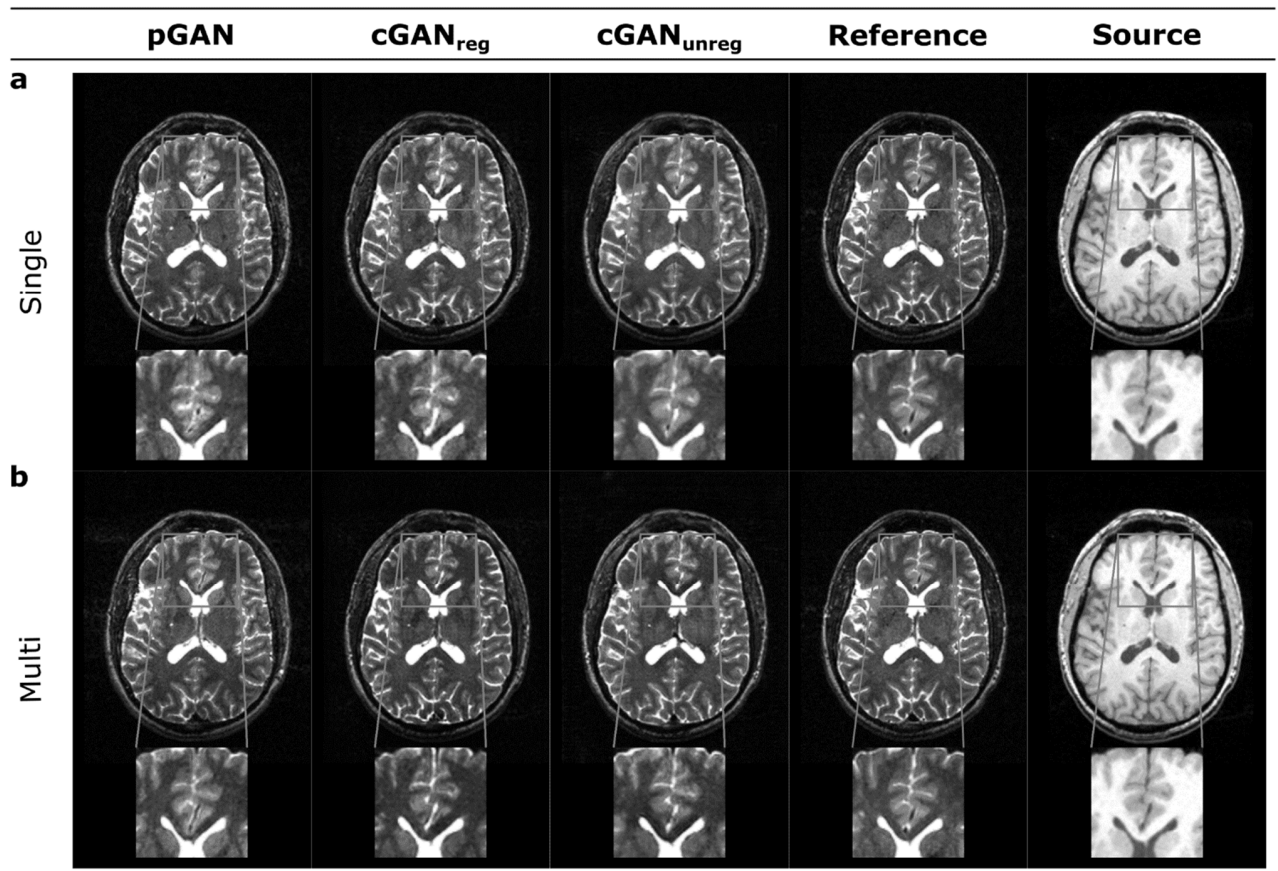

**Figure 3.** The proposed approach was demonstrated for synthesis of $T_2$-weighted images from $T_1$-weighted images in the MIDAS dataset. Synthesis was performed with pGAN, cGAN trained on registered images (cGAN$_{reg}$), and cGAN trained on unregistered images (cGAN$_{unreg}$). For pGAN and cGAN$_{reg}$, training was performed using $T_2$-weighted images registered onto $T_1$-weighted images ($T_1 \rightarrow T_2^*$). Synthesis results for (a) the single cross-section, and (b) multi cross-section models are shown along with the true target image (reference) and the source image (source). Zoomed-in portions of the images are also displayed. While both pGAN and cGAN yield synthetic images of striking visual similarity to the reference, pGAN is the top performer. Synthesis quality is improved as information across neighboring cross sections is incorporated, particularly for the pGAN method.

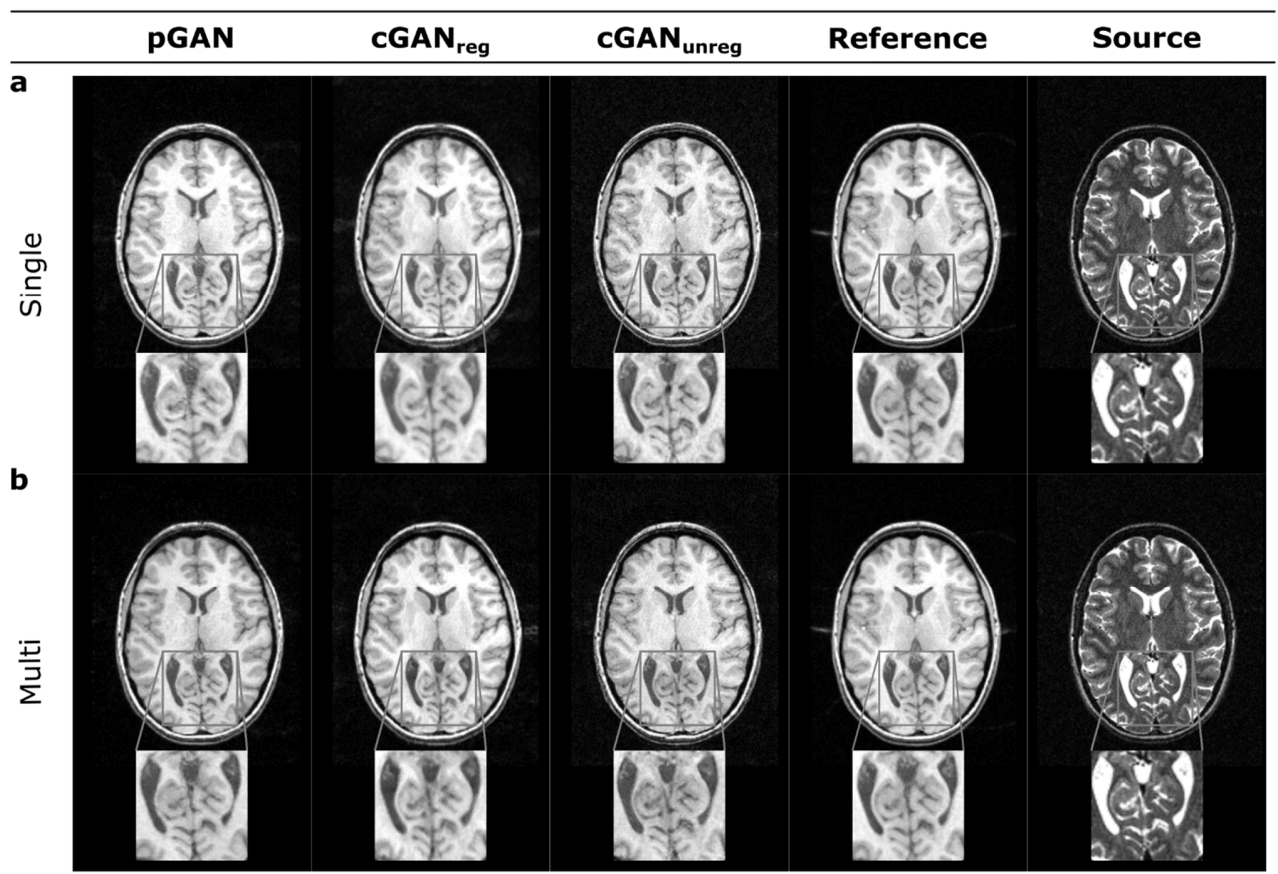

**Figure 4.** The proposed approach was demonstrated for synthesis of $T_1$-weighted images from $T_2$-weighted images in the MIDAS dataset. Synthesis was performed with pGAN, cGAN trained on registered images (cGAN$_{reg}$), and cGAN trained on unregistered images (cGAN$_{unreg}$). For pGAN and cGAN$_{reg}$, training was performed using $T_1$-weighted images registered onto $T_2$-weighted images ($T_2 \rightarrow T_1$*). Synthesis results for (a) the single cross-section, and (b) multi cross-section models are shown along with the true target image (reference) and the source image (source). Zoomed-in portions of the images are also displayed. While both pGAN and cGAN yield synthetic images of striking visual similarity to the reference, pGAN is the top performer. Synthesis quality is improved as information across neighboring cross sections is incorporated, particularly for the pGAN method.

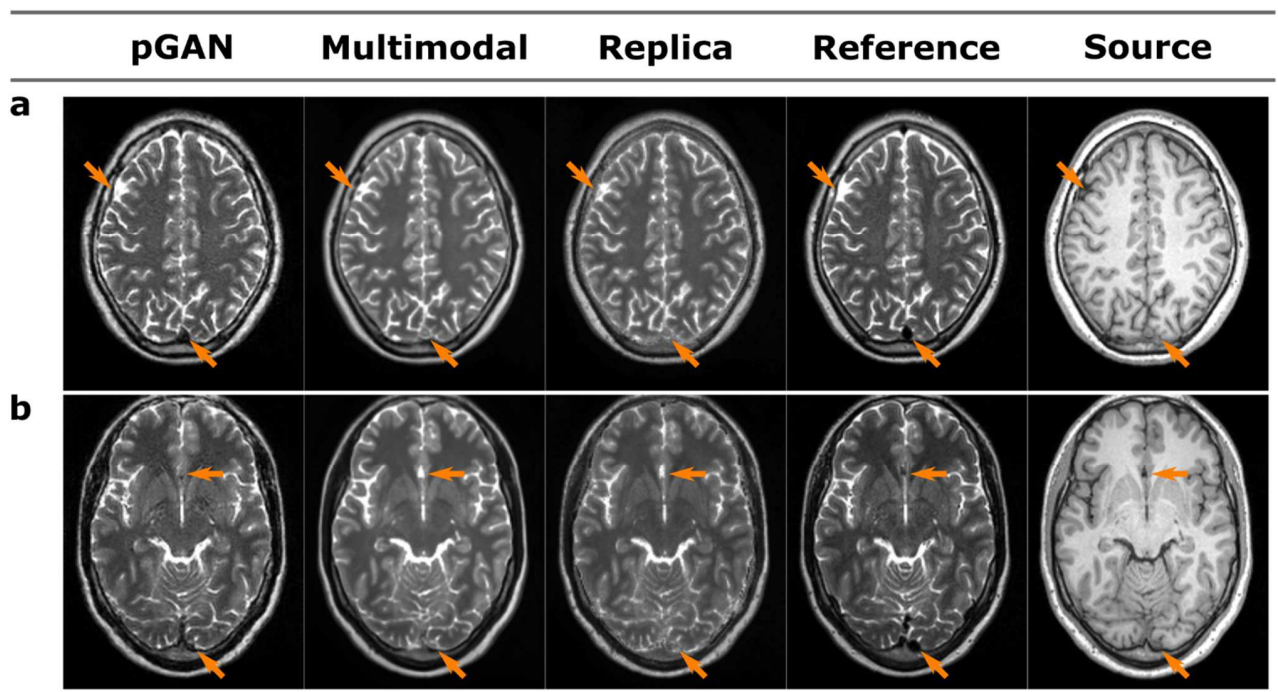

**Figure 5.** The proposed approach was demonstrated for synthesis of $T_2$-weighted images from $T_1$-weighted images in the IXI dataset. $T_1 \rightarrow T_2$* synthesis was performed with pGAN, Multimodal and Replica. Synthesis results are shown for a sample cross section from two different subjects (a and b) along with the true target image (reference) and the source image (source). The proposed method outperforms competing methods in terms of synthesis quality. Regions near issue boundaries that are inaccurately synthesized by the competing methods are reliably depicted by pGAN (marked with arrows). The use of adversarial loss functions enables improved accuracy in synthesis of high-spatial-frequency texture in $T_2$-weighted images compared to Multimodal and Replica that show some degree of blurring.

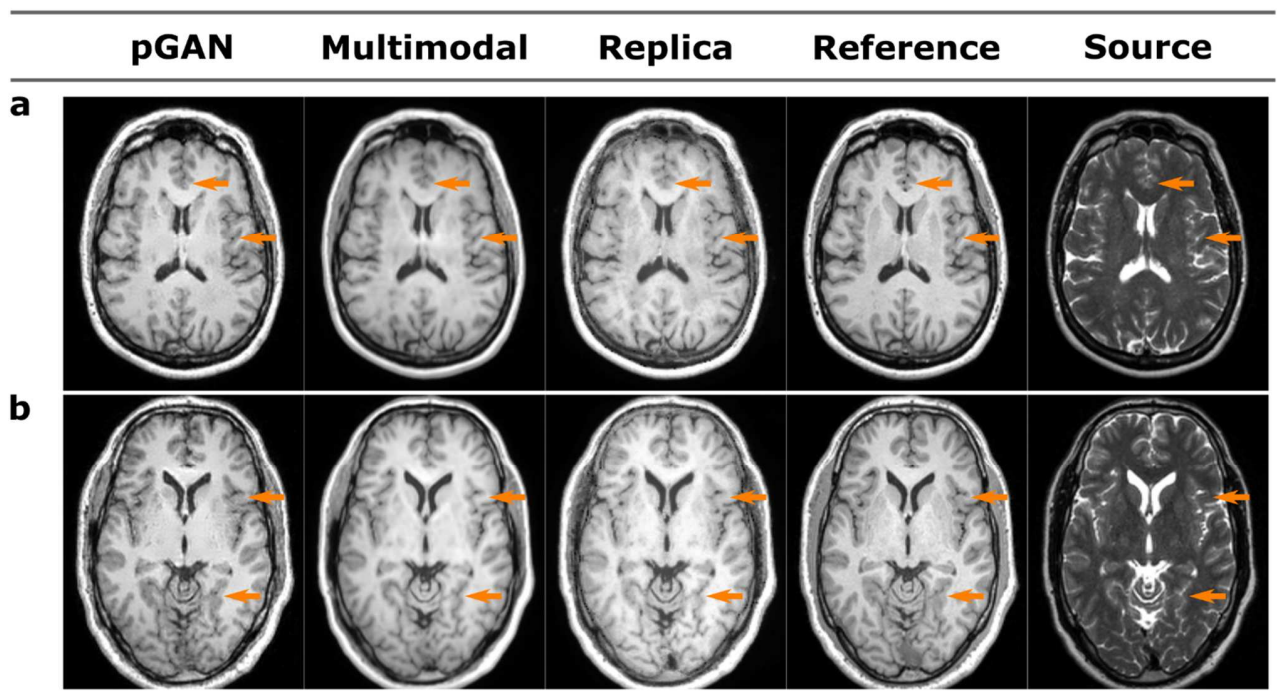

**Figure 6.** The proposed approach was demonstrated for synthesis of $T_1$-weighted images from $T_2$-weighted images in the IXI dataset. $T_2^* \rightarrow T_1$ synthesis was performed with pGAN, Multimodal and Replica. Synthesis results are shown for a sample cross section from two different subjects (a and b) along with the true target image (reference) and the source image (source). The proposed method outperforms competing methods in terms of synthesis quality. Regions near issue boundaries that are inaccurately synthesized by the competing methods are reliably depicted by pGAN (marked with arrows). The use of adversarial loss functions enables improved accuracy in synthesis of high-spatial-frequency texture in $T_2$-weighted images compared to Multimodal and Replica that show some degree of blurring.

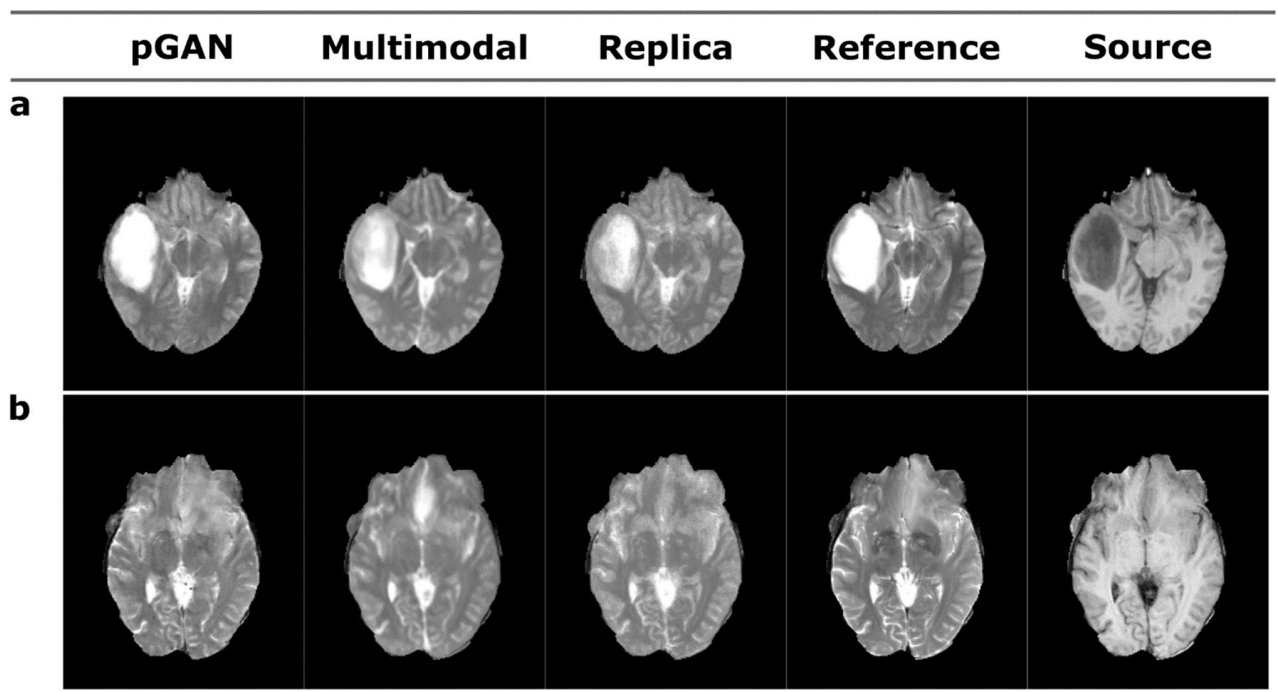

**Figure 7.** The proposed approach was demonstrated on glioma patients for synthesis of $T_2$-weighted images from $T_1$-weighted images in the BRATS dataset. $T_1 \rightarrow T_2^*$ synthesis was performed with pGAN, Multimodal and Replica. Synthesis results are shown for a sample cross section from two different subjects (a and b) along with the true target image (reference) and the source image (source). Certain pathologies present in the source contrast lead to artefactual synthesis with Replica and Multimodal (marked with arrows). Meanwhile, the pGAN method enables reliable synthesis with visibly improved depiction of high spatial frequency information.

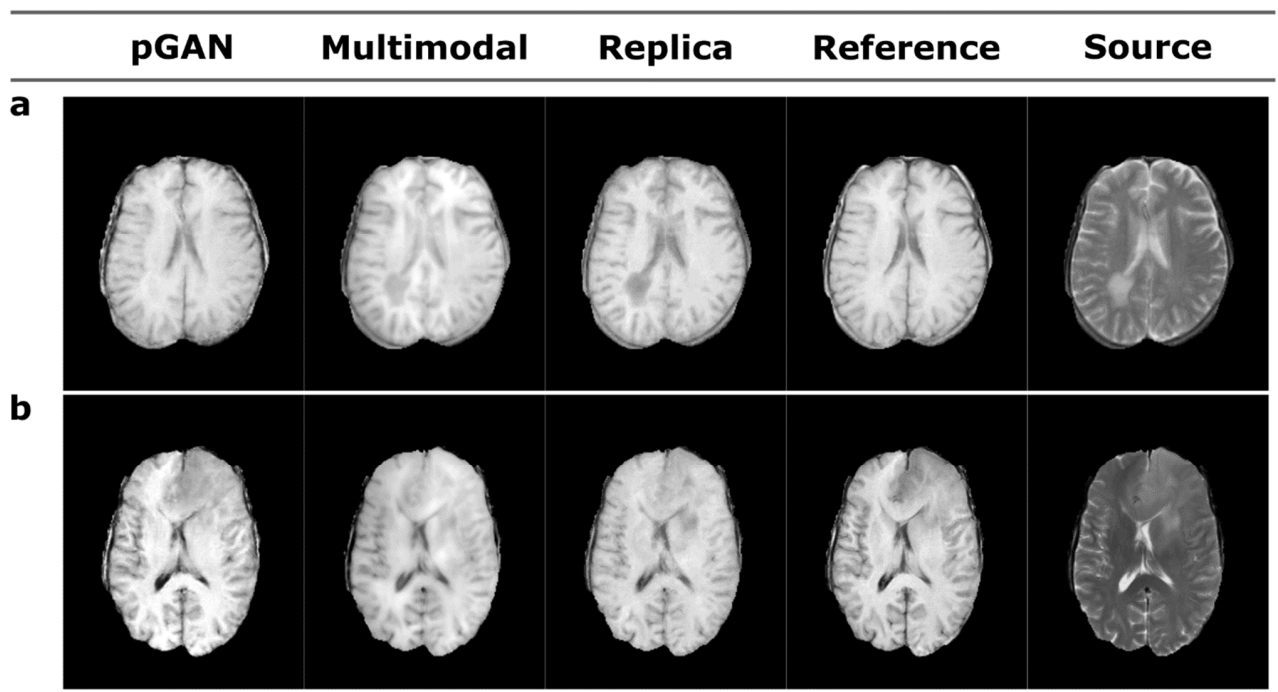

**Figure 8.** The proposed approach was demonstrated on glioma patients for synthesis of $T_1$-weighted images from $T_2$-weighted images in the BRATS dataset. $T_2^* \rightarrow T_1$ synthesis was performed with pGAN, Multimodal and Replica. Synthesis results are shown for a sample cross section from two different subjects (a and b) along with the true target image (reference) and the source image (source). Certain pathologies present in the source contrast lead to artefactual synthesis with Replica and Multimodal (marked with arrows). Meanwhile, the pGAN method enables reliable synthesis with visibly improved depiction of high spatial frequency information.

# TABLES

**Table I.** Quality of synthesis in the MIDAS dataset. Synthesis performance for $T_1$- and $T_2$-weighted images was evaluated for cGAN trained on unregistered images (cGAN$_{unreg}$), cGAN trained on registered images (cGAN$_{reg}$), and pGAN. Two separate models were tested for synthesis of $T_2$-weighted contrast ($T_1 \rightarrow T_2^*$, $T_1^* \rightarrow T_2$), and for synthesis of $T_1$-weighted contrast ($T_2 \rightarrow T_1^*$ $T_2^* \rightarrow T_1$). $T_1^*$ is an image registered onto the respective $T_2$, and $T_2^*$ is an image registered onto the respective $T_1$. SSIM and PSNR measurements are reported as mean±std across test images. For each synthesis task, the highest measurements for PSNR and SSIM are marked in bold font.

|  | cGAN$_{unreg}$ | | cGAN$_{reg}$ | | pGAN | |
|---|---|---|---|---|---|---|
|  | SSIM | PSNR | SSIM | PSNR | SSIM | PSNR |
| $T_1 \rightarrow T_2^*$ | 0.756 ± .039 | 20.94 ± 1.10 | 0.840 ± .051 | 22.03 ± 1.67 | **0.878 ± .046** | **24.93 ± 1.60** |
| $T_1^* \rightarrow T_2$ | 0.743 ± .030 | 20.07 ± 1.16 | 0.765 ± .040 | 20.75 ± 1.03 | **0.818 ± .044** | **23.28 ± 1.45** |
| $T_2 \rightarrow T_1^*$ | 0.788 ± .034 | 23.59 ± 1.98 | 0.822 ± .038 | 24.09 ± 2.41 | **0.880 ± .027** | **25.55 ± 2.37** |
| $T_2^* \rightarrow T_1$ | 0.781 ± .033 | 23.49 ± 1.60 | 0.853 ± .029 | 24.80 ± 1.81 | **0.854 ± .036** | **24.45 ± 2.24** |

**Table II.** Quality of synthesis in the MIDAS dataset. Synthesis performance for $T_1$- and $T_2$-weighted images was evaluated for cGAN$_{unreg}$, cGAN$_{reg}$, and pGAN. Two separate models were tested for synthesis of $T_2$-weighted contrast ($T_1 \rightarrow T_2^*$, $T_1^* \rightarrow T_2$), and for synthesis of $T_1$-weighted contrast ($T_2 \rightarrow T_1^*$ $T_2^* \rightarrow T_1$). Synthesis was performed on three consecutive cross-sections. SSIM and PSNR measurements are reported as mean±std across test images. For each synthesis task, the highest measurements for PSNR and SSIM are marked in bold font.

|  | cGAN$_{unreg}$ | | cGAN$_{reg}$ | | pGAN | |
|---|---|---|---|---|---|---|
|  | SSIM | PSNR | SSIM | PSNR | SSIM | PSNR |
| $T_1 \rightarrow T_2^*$ | 0.763 ± .048 | 21.43 ± 0.99 | 0.832 ± .050 | 22.27 ± 1.79 | **0.897 ± .043** | **25.74 ± 1.59** |
| $T_1^* \rightarrow T_2$ | 0.762 ± .052 | 20.71 ± 1.10 | 0.770 ± .059 | 20.55 ± 1.58 | **0.825 ± .046** | **23.56 ± 1.52** |
| $T_2 \rightarrow T_1^*$ | 0.790 ± .044 | 22.70 ± 1.60 | 0.874 ± .032 | 24.42 ± 2.51 | **0.904 ± .025** | **26.31 ± 2.68** |
| $T_2^* \rightarrow T_1$ | 0.770 ± .052 | 21.47 ± 1.73 | 0.797 ± .039 | 23. 78 ± 1.56 | **0.880 ± .031** | **25.67 ± 2.03** |

**Table III.** Method comparison in the IXI dataset. Synthesis performance for $T_1$- and $T_2$-weighted images was evaluated for cGAN$_{reg}$, pGAN, Replica and Multimodal. Two separate models were tested for synthesis of $T_2$-weighted contrast ($T_1 \rightarrow T_2^*$, $T_1^* \rightarrow T_2$), and for synthesis of $T_1$-weighted contrast ($T_2 \rightarrow T_1^*$ $T_2^* \rightarrow T_1$). SSIM and PSNR measurements are reported as mean±std across test images. For each synthesis task, the highest measurements for PSNR and SSIM are marked in bold font.

|  | cGAN$_{reg}$ | | pGAN | | Replica | | Multimodal | |
|---|---|---|---|---|---|---|---|---|
|  | SSIM | PSNR | SSIM | PSNR | SSIM | PSNR | SSIM | PSNR |
| $T_1 \rightarrow T_2^*$ | 0.934±.025 | 27.37±1.33 | **0.936±.022** | **27.51±1.36** | 0.906±.036 | 24.75±2.22 | 0.932±.024 | 26.70±1.40 |
| $T_1^* \rightarrow T_2$ | 0.900±.036 | 25.76±1.40 | **0.907±.035** | **26.32±1.56** | 0.859±.048 | 23.31±2.18 | 0.893±.035 | 25.06±1.58 |
| $T_2 \rightarrow T_1^*$ | 0.907±.040 | 25.29±1.91 | **0.916±.037** | **25.81±1.87** | 0.866±.056 | 20.72±2.22 | 0.907±.041 | 23.64±2.22 |
| $T_2^* \rightarrow T_1$ | 0.941±.027 | 27.33±1.82 | **0.944±.025** | **27.37±1.78** | 0.901±.042 | 22.02±2.38 | 0.936±.029 | 24.71±2.06 |

**Table IV.** Method comparison in the BRATS dataset. Synthesis performance for $T_1$- and $T_2$-weighted images was evaluated for cGAN$_{reg}$, pGAN, Replica and Multimodal. Two separate models were tested for synthesis of $T_2$-weighted contrast ($T_1 \rightarrow T_2^*$, $T_1^* \rightarrow T_2$), and for synthesis of $T_1$-weighted contrast ($T_2 \rightarrow T_1^*$ $T_2^* \rightarrow T_1$). SSIM and PSNR measurements are reported as mean±std across test images. For each synthesis task, the highest measurements for PSNR and SSIM are marked in bold font.

|  | cGAN$_{reg}$ |  | pGAN |  | Replica |  | Multimodal |  |
|---|---|---|---|---|---|---|---|---|
|  | SSIM | PSNR | SSIM | PSNR | SSIM | PSNR | SSIM | PSNR |
| $T_1 \rightarrow T_2^*$ | 0.943±.017 | 25.82±1.94 | **0.955±.017** | **26.95±2.62** | 0.952±.018 | 25.30±2.60 | 0.948±.018 | 24.61±2.37 |
| $T_2^* \rightarrow T_1$ | 0.945±.017 | 25.72±2.09 | **0.956±.013** | **26.96±2.49** | 0.949±.020 | 24.43±4.81 | 0.954±.017 | 24.69±3.13 |